\documentclass[sigconf,nonacm]{acmart}
\setcopyright{none}
\renewcommand\footnotetextcopyrightpermission[1]{}
\AtBeginDocument{%
  }

\usepackage{booktabs}  % nice table rules
\usepackage{multirow}  % enables \multirow
\usepackage{tablefootnote} % in preamble
\usepackage{algorithm} 
\usepackage{algpseudocode}
\usepackage{makecell}
\usepackage{enumitem}
\usepackage{float}
\begin{document}

\title{TabChange: Precise Attribute Changes in Tabular Data}

\author{Arjun Dahal}
\affiliation{%
  \institution{The University of Texas at Arlington}
  \city{Arlington}
  \state{Texas}
  \country{USA}}
\email{axd5000@mavs.uta.edu}

\author{Yu Lei}
\affiliation{%
  \institution{The University of Texas at Arlington}
  \city{Arlington}
  \state{Texas}
  \country{USA}}
\email{ylei@cse.uta.edu}

\author{Raghu N. Kacker}
\affiliation{%
  \institution{Information Technology Laboratory,\\National Institute of Standards and Technology}
  \city{Gaithersburg}
  \state{Maryland}
  \country{USA}}
\email{raghu.kacker@nist.gov}

\author{Richard Kuhn}
\affiliation{%
  \institution{Information Technology Laboratory,\\
    National Institute of Standards and Technology}
  \city{Gaithersburg}
  \state{Maryland}
  \country{USA}}
\email{d.kuhn@nist.gov}

\begin{abstract}
Modifying an attribute in tabular data often introduces an unnatural instance by breaking its relationships with other attributes. The modified instance must be both natural and minimally changed from the original instance. This paper addresses the challenge of generating such a modified instance. We identify key limitations in existing approaches: generative models either don't support instance-level attribute editing or, in the case of methods like CVAE, retain attribute information in the latent space, leading to unnecessary modifications. To solve this, we propose TabChange, an approach that analyzes the relationship between the attribute of interest and other attributes in the dataset. If the relationship is weak, it simply flips the attribute; if it is strong, it uses an adversarial framework that removes information about the attribute in the latent space representation. This removal enables precise modifications, making only the necessary adjustments to maintain naturalness. Our experiments across seven datasets show that TabChange generates counterfactuals in attributes that are comparable in naturalness and are more proximal to their original instances. This leads to a higher number of valid counterfactuals and a lower number of invalid counterfactuals compared to the baselines.
\end{abstract}



%%
%% This command processes the author, affiliation, and title
\maketitle

\section{Introduction}

Machine Learning systems, particularly generative models, have become common in processing unstructured data such as images and text. However, high-stakes decisions ranging from loan approval \cite{mukerjee2002multi} to healthcare diagnostics are based on tabular data. In these domains, the challenge extends beyond prediction to attribute modification, where specific attributes of existing instances need to be changed. The essential criteria to change an attribute of an instance are: the identity of the instance should be preserved, and the change shouldn't break the relationship of the attribute with other attributes. For example, when changing the gender of an individual having Executive Officer as an occupation, the changed instance should also have a similar occupation representing education and ambition that was present in the original instance. Simply flipping an attribute might break the relationship with related attributes. For instance, changing an individual’s occupation from Engineer to Nurse would likely require changes in attributes like average work hours and income. Ignoring dependencies between the attributes during the change results in an unnatural instance.

Existing generative models, such as Diffusion models and Generative Adversarial Networks, are suitable for generation, but not for changing an attribute of a specific tabular instance. These models generate instances from noise based on conditioning, but do not have a mechanism to preserve the identity of the original instance, which is essential for attribute modification. On the other hand, CVAEs encode an instance in the latent space and can decode conditioned on different values of the attribute, but the latent space still carries information about the previous value of the attribute. This leads to unnecessary changes in the generated instance when conditioning on a different attribute value. Thus, existing generative approaches either lose the identity of the instance or result in excessive modifications.

Current methods for counterfactual generation focus on changing prediction labels and optimizing for validity (prediction flip), proximity, sparsity, and diversity. These approaches involve search-based approaches \cite{mothilal2020explaining} \cite{ustun2019actionable} \cite{poyiadzi2020face}, or generative  \cite{joshi2019towards} \cite{pawelczyk2020learning} \cite{madaan2024navigating} \cite{panagiotou2024tabcf} \cite{garg2025search} \cite{downs2020cruds} that aim to find the smallest change in attributes that changes the model's prediction. The attribute changes are guided by the model's internal logic, rather than the data distribution. In contrast, our work focuses on attribute-level changes that are natural, i.e., satisfy data constraints and proximity.

In fairness testing \cite{saleiro2018aequitas} \cite{zhang2021automatic} \cite{zhang2021efficient} \cite{zheng2022neuronfair} \cite{fan2022explanation} \cite{xiao2023latent} \cite{yin2024boundary}, existing approaches simply flip the value of the sensitive attribute (e.g., \textit{race}, \textit{age}, \textit{sex}) to generate counterfactual pairs for testing machine learning model predictions. However, this ignores the dependencies between the sensitive attribute and other attributes.
Causal approaches \cite{cinquini2024causality} \cite{van2021decaf} use the Structural Causal Model (SCMs) to generate natural instances by explicitly modeling attribute dependencies. However, they are often impractical because while SCMs can be constructed automatically, this process is error-prone and often requires human expertise to validate and correct the model.

The current approach for generating counterfactuals on an attribute is CVAE \cite{sohn2015learning}, which is used by \cite{majumdar2021counterfactuals} to generate counterfactuals on sensitive attributes by conditioning the decoder with a different attribute value during inference. However, since the latent space is not independent of the attribute being changed, this leads to unnecessary modifications. Recently, diffusion models have shown promise in generating tabular data, but the instance is converted to noise in the latent space before the reverse diffusion process is applied. Since the identity of the instance is lost in the latent space, additional changes are required for counterfactual generation.

Thus, generating instances that meet the strict requirements of naturalness and proximity remains challenging. Tabular data typically includes a mix of discrete and continuous variables and can be highly imbalanced, making it difficult to model the data distribution accurately and change attribute values in a precise, controlled manner. For any instance with a changed attribute, the generated instance should be natural, i.e. preserves relationship between attributes so it can exist in the real world, and proximal, meaning minimally changed from the original. For fairness testing to be valid, any change in the model's outcome must be attributable to the change in the sensitive attribute, requiring only necessary adjustments to related attributes without introducing unrelated modifications. In the rest of the paper, we refer to counterfactuals in attributes simply as counterfactuals unless otherwise indicated.

We propose TabChange, a novel approach for changing attributes in tabular data. Our approach is based on two key ideas. First, we analyze the relationship between the attribute of interest and other attributes in the dataset to select the appropriate modification technique—either a simple attribute flip or an adversarial framework. If the attribute of interest has a weak relationship with other attributes, we simply flip the attribute. Otherwise, we train an encoder-decoder network along with a discriminator on two competing objectives: reconstruction and adversarial training to create a latent space that doesn't contain information about the attribute. This allows for controlled changes to generate counterfactuals by conditioning the decoder while maintaining naturalness.

We evaluate TabChange on six benchmark datasets and one synthetic dataset, comparing it against CVAE and Diffusion baselines. Our experiments measured the ratio of natural counterfactual instances retained and proximity (average distance of retained counterfactuals to their original instances). We combine these metrics into a single unified metric, Valid Counterfactual Rate (VCR), which measures the percentage of counterfactuals that are both natural and more proximal than competing methods. The results show that TabChange achieves higher Valid Counterfactual Rates in single-attribute cases and multi-attribute cases compared to the baselines.\\
Our major research contributions are as follows:
\begin{enumerate}
\item \textbf{Approach}: We propose a novel two-stage approach for changing attribute values in tabular data. We first analyze attribute relationships to adaptively select between simple attribute flipping and adversarial training for attribute-invariant latent space construction, leading to counterfactuals that are both natural and more proximal to the original instances.

\item \textbf{Evaluation}: We performed an experimental evaluation of our approach across seven datasets using Valid Counterfactual Rate as our primary metric. The results indicate that TabChange achieves higher VCR in 8 out of 11 single-attribute cases and 3 out of 4 multi-attribute cases compared to the baselines.

\item \textbf{Tool}: We implemented our approach in a publicly available research prototype tool\footnote{\url{https://anonymous.4open.science/r/TabChange-409E/}}.
\end{enumerate}

\section{Related Work:}
We focus on the existing works that tackle the problem of changing the attribute of the tabular data while making appropriate changes to other attributes related to the changed attribute. The problem of changing a feature has been widely tackled in the image domain \cite{kim2021counterfactual} \cite{perarnau2016invertible} \cite{nemirovsky2020countergan} \cite{lample2017fader}. However, this has not been adequately addressed in the tabular domain. Unlike images, which consist of pixel values along with spatial dependencies, tabular data presents unique complexities due to a mix of continuous and categorical attributes and the relationships between multiple attributes. Furthermore, the values of an attribute might be imbalanced. This requires us to change the value of related attributes when the attribute of interest is changed. Our task is different from \cite{rajabi2022tabfairgan} \cite{xu2019fairgan+} \cite{van2021decaf} \cite{yang2024balanced}, which focus on generating an entire synthetic tabular dataset that adheres to a group fairness metric. The approaches for changing attribute values can be broadly categorized into simple flipping, causal modeling, and learning-based approaches.

The most straightforward approach to change the attribute of an instance exists in fairness testing literature, such as \cite{saleiro2018aequitas} \cite{zhang2021automatic} \cite{zhang2021efficient} \cite{zheng2022neuronfair} \cite{fan2022explanation}, which involves simply changing the value of a sensitive attribute from a previleged group to an unprevileged group and vice-versa to create a test pair on sensitive attribute. While this technique is simple to implement for generating test pairs, it disregards the naturalness of the instance flipped on the sensitive attribute. Unlike previous works, 
\cite{xiao2023latent} \cite{yin2024boundary} focus on generating natural instances for fairness testing. However, they also simply flip to obtain a test pair for fairness testing. These approaches do not adjust related attributes related to the sensitive attribute, which can undermine the validity of the fairness testing. In contrast, our approach is designed to produce counterfactuals that are both natural and minimally changed.

Causal approaches \cite{van2021decaf} \cite{cinquini2024causality} rely on a formal Structural Causal Model (SCM) that explicitly defines the cause-and-effect relationships between attributes for their task. SCMs might be used to explain \cite{cinquini2024causality} or generate fair data \cite{van2021decaf}. With a known SCM, a counterfactual can be computed in a way that is guaranteed to respect the defined causal structure. A causal graph can be automatically constructed using causal discovery algorithms; however, this process could result in the identification of wrong cause-and-effect relationships. Furthermore, a discovered causal graph often requires validation from a human expert, requiring manual effort to process, and thus doesn’t scale to new datasets. Our approach overcomes this limitation by learning the data distribution directly.

Counterfactual generation approaches find counterfactuals for a machine learning model's decision. Search-based approaches \cite{mothilal2020explaining} \cite{ustun2019actionable} \cite{poyiadzi2020face} find counterfactuals for a different decision of a Machine Learning model by formulating it as a search problem. These approaches aim to find a new instance while minimizing a distance such as L1 (encouraging sparsity) or L2 distance to the original instance. To ensure naturalness, they rely on user-defined constraints, such as defining which attributes are immutable or specifying valid ranges. While effective for simple changes, these approaches struggle to capture complex dependencies between attributes unless those relationships are explicitly encoded as constraints, which is not scalable. Some approaches \cite {joshi2019towards} \cite{pawelczyk2020learning} use VAE to find counterfactuals that change the prediction of a machine learning model. Their process involves an optimization that searches the latent space for a change that will change the machine learning model's decision. The diffusion model is used in \cite{madaan2024navigating} to generate counterfactuals, applying a guided denoising process to construct a valid new instance. Transformer-based generative models are leveraged by \cite{panagiotou2024tabcf} \cite{garg2025search} to model complex dependencies in tabular data. \cite{downs2020cruds} produces instances that respect dependencies in the data and apply causal and individualized constraints. Unlike approaches that depend on producing counterfactuals for a machine learning model's decision, \cite{majumdar2021counterfactuals} uses CVAE to generate counterfactuals on sensitive attributes for fairness testing. However, the latent space is generated under the influence of the original attribute, not independent of it. Thus, the counterfactual might not be proximal to the original instance. All of these approaches except \cite{majumdar2021counterfactuals} focus on generating instances for the purpose of changing a model's prediction, but do not target change based only on the data distribution, which is central to our approach. Unlike \cite{majumdar2021counterfactuals}, our approach first decides whether to simply flip or use an adversarial framework to generate attribute-invariant latent space and combines it with a different value of the attribute to obtain counterfactuals with minimal changes. This improves efficiency by not using a generative model when it's not required, i.e., when the attribute of interest has a weak relationship with other attributes.

\section{Background}
In this section, we review the concept of Latent Space Invariance, which is central to our approach for generating attribute-invariant representations.

\subsection{Latent Space Invariance}
The goal of learning an invariant representation is to map an input $x$ to a latent vector $z$ such that $z$ retains information about $x$ while being independent of a specific attribute. This requires the distribution of latent codes to be identical across different attribute values by minimizing the information between the attribute and the representation. There are two primary approaches to achieving this invariance: Information-Theoretic Invariance and Adversarial Training.

\subsubsection{Information-Theoretic Invariance}
These approaches \cite{zemel2013learning} \cite{louizos2015variational} impose invariance by penalizing the distance between the latent distributions of different attribute values. The discrepancy between probability distributions for different values of the attribute is measured as a statistical metric and used as a regularizer to the loss function. \cite{zemel2013learning} uses statistical parity, whereas \cite{louizos2015variational} uses Maximum Mean Discrepancy to remove group-specific information,n making the latent representation indistinguishable to downstream classifiers without requiring an adversary.

\subsubsection{Adversarial Training}
These methods \cite{edwards2015censoring} \cite{lample2017fader} \cite{madras2018learning} replace fixed statistical penalties with a min-max game between two competing neural networks with often conflicting objectives to learn attribute invariant latent space. Instead of minimizing a metric, these methods employ a separate neural network, a discriminator, that attempts to predict the attribute $a$ from the latent code $z$. The training process is formulated as a zero-sum game, often using an autoencoder network: the discriminator is optimized to maximize its accuracy in predicting $a$, while the encoder is trained adversarially to fool the discriminator, and the decoder minimizes reconstruction loss. By optimizing for both reconstruction and this adversarial objective, the encoder learns to remove the attribute's information from the latent representation, forcing the decoder to rely on the user-provided attribute value for accurate reconstruction.

In our work, we employ adversarial training to achieve latent space invariance when the attribute of interest has a strong relationship with other attributes. By removing attribute information from the latent representation, we can then condition the decoder on a different attribute value, ensuring that only the necessary adjustments are made to related attributes while preserving the instance's identity.

\section{Approach}
\subsection{Problem Setup}

\label{subsec:problem-setup}

Let $x = (x_{con}, x_{cat}) \in \mathcal{X}$ denote a tabular instance, where $x_{con}$ and $x_{cat}$ represent the subsets of continuous and categorical attributes, indexed by sets $C$ and $G$ respectively. Let $a$ be the attribute of
interest, and let $v$ be its observed value. We write
\begin{equation}
x = (x_{\setminus a},\, a=v)
\end{equation}
where $x_{\setminus a}$ denotes all features except the attribute of interest $a$. Given a value $v'\neq v$ of the attribute, our goal is to produce a counterfactual instance $\hat{x}$ on attribute $a$.
\begin{equation}
\hat{x} = (\hat{x}_{\setminus a},\, a=v')
\end{equation}

\noindent The counterfactual instance $\hat{x}$ must be minimally changed from $x$ and natural by satisfying the underlying constraints of the dataset.
\subsection{Overview of Our Approach}
\label{subsec:overview}

Our key observation is that changes to be made in an instance depend on how strongly the attribute of interest is related to other attributes. If the attribute has a weak relationship with other attributes, then no adjustments to related attributes are necessary. In such cases, simply changing the target attribute is sufficient and more efficient than training a generative model. Conversely, when the attribute of interest is strongly related to other columns, simply flipping breaks the relationship with other attributes; we need to remove information about the old value of the attribute before adding the new value of the attribute so that the related attributes are adjusted in a controlled way.

Our approach, TabChange, first measures the Mutual Information (MI) between the attribute of interest and the remaining attributes, then adapts the strategy accordingly. When the MI is below a threshold, we directly flip the attribute. When it is above the threshold, we train an adversarial framework that learns an attribute-invariant latent space and conditions the decoder on the new attribute value. This MI-guided process enables minimal changes to the instance while preserving naturalness.\\
\noindent TabChange has two main steps:

\begin{enumerate} [leftmargin=*]
\item {Attribute Relationship Analysis:}
  We estimate how strongly the attribute of interest $a$ is related to each of the remaining attributes $x_{\setminus a}$
  using mutual information. 

\item {Counterfactual Generation:}
  Given a value of the attribute of interest $v'\neq v$, we generate a counterfactual instance 
  $\hat{x}=(\hat{x}_{\setminus a}, a=v')$
  using one of the following strategies based on attribute relationship analysis. 
  \begin{enumerate}
    \item {Direct flip:} If the relationship is weak, we simply flip the attribute as $a=v'$ and keep the remaining attributes unchanged.
    \item {Adversarial Framework:} If the relationship between the attribute of interest and other attributes is strong, we pass $x$ in the adversarial framework to obtain attribute-invariant representation in the latent space and decode it conditioned on $a=v'$.
  \end{enumerate}
\end{enumerate}

\noindent The extension to editing multiple attributes is discussed in Subsection~\ref{subsec:multiple-attribute-changes}.

\begin{figure}[t]
  \centering
  \includegraphics[width=0.85\linewidth]{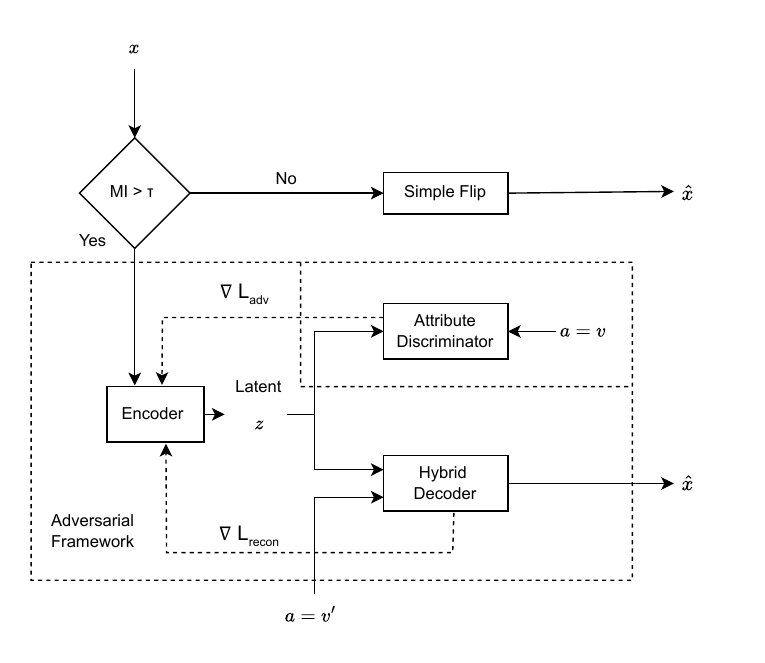} 
  \caption{TabChange first determines the MI of the attribute of interest with other attributes. If the relationship is weak, the attribute is simply flipped. For (MI $> \tau$), a strong relationship, the instance is passed through an adversarial framework that creates an attribute invariant latent space before decoding with the new value of the attribute.}
  \label{fig:framework}
\end{figure}

\subsection{Attribute Relationship Analysis}
TabChange first analyzes how strongly the attribute of interest $a$ is related to each of the remaining
attributes $x_{\setminus a}$ in the dataset. 
To measure the strength of this relationship, we compute mutual information between the attribute of interest $a$ and
other attributes.
If mutual information exceeds a threshold $\tau$, we treat the relationship as strong; otherwise, we treat it as weak.

\subsection{Counterfactual Generation}
Given an instance $x=(x_{\setminus a}, a=v)$ and a value of attribute of interest $v'\neq v$, our goal is to generate a
counterfactual instance on the attribute $a$,
\begin{equation}
  \hat{x}=(\hat{x}_{\setminus a}, a=v'). 
\end{equation}
TabChange generates $\hat{x}$ using one of two strategies, selected by calculating mutual information between the attribute of interest and other attributes.

\subsubsection{Direct flip:}
If the relationship is weak (MI $\le \tau$), we directly set $a=v'$ and keep the remaining attributes
unchanged, i.e., $\hat{x}_{\setminus a}=x_{\setminus a}$.

\subsubsection{Adversarial Framework:}
\label{subsec:adv-framework}
If the relationship is strong \(\mathrm{MI} > \tau\), we use an adversarial framework to learn a latent
representation that removes information about the attribute of interest $a$.
At inference time, we obtain an attribute-invariant representation $z$ and decode it conditioned on the
value $v'$ to generate $\hat{x}$, making only the necessary adjustments to the remaining attributes.

The adversarial framework employs a competing training dynamic where the discriminator tries to predict the attribute from the latent code while the encoder prevents this prediction by removing attribute-related information. This ensures that when we decode with a new attribute value, only the necessary related attributes are adjusted.

We propose an adversarial framework with encoder $E_{\theta_{\text{enc}}}$, decoder $D_{\theta_{\text{dec}}}$, and discriminator $D_{\theta_{\text{dis}}}$ in the latent space. The encoder and decoders have fully connected layers with ReLU activations\cite{xu2019modeling}, and the discriminator uses Leaky ReLU activation to ensure it provides a consistent and effective gradient signal to the encoder throughout the training process. To handle a mix of continuous and categorical attributes, we employ a hybrid reconstruction loss function in the decoder. We use Mean Squared Error (MSE) for the continuous attributes and Cross-Entropy loss for the categorical attributes \cite{patki2016synthetic}. Categorical attributes in real-world datasets are often highly imbalanced. To prevent the discriminator from simply guessing the majority class, we introduce a class-weighted loss function in the discriminator.
Our training process emphasizes reconstruction for the first few epochs, then gradually increases the adversarial weight and caps it, while alternating training between the discriminator and encoder-decoder till the training converges.

For a training instance $x_i \in \mathcal{D}$ with its latent space representation $z_i$, the discriminator attempts to predict $a_i$ from $z_i$. Let $x'_i$ denote the instance reconstructed by the decoder. To ensure the model preserves the original features, the reconstruction loss is defined as:

\begin{equation}
    \mathcal{L}_{recon} = \sum_{j \in C \setminus \{a\}} (x'_{i,j} - x_{i,j})^2 + \sum_{g \in G \setminus \{a\}} (-y^{(g)\top} \log p^{(g)})
\end{equation}

\noindent where the first term is the Mean Squared Error (MSE) for continuous attributes and the second term is the cross-entropy loss for categorical attributes.

\noindent The discriminator is trained using a class-weighted cross-entropy loss over a batch $\mathcal{B}$:
\begin{equation}
    \mathcal{L}_{dis} = \frac{1}{|\mathcal{B}|} \sum_{(x_i, a_i) \in \mathcal{B}} \left[ -\alpha_{a} \left( \text{softmax}(D_{\theta_{dis}}(z_i)) \right) \right],\quad z_i = E_{\theta_{enc}}(x_i)
\end{equation}

where $|\mathcal{B}|$ denotes the batch size, and $\alpha_a$ is a class-specific weight used to correct for imbalance in the values of $a$. To enforce invariance of the latent representation $z$ with respect to $a$, 
The encoder is trained adversarially against the discriminator. 
While the discriminator minimizes $\mathcal{L}_{\mathrm{dis}}$ to recover $a$ from $z$, 
The encoder is updated with the opposite objective to produce attribute invariant latent representation.

After training, we generate a counterfactual $\hat{x}$ by encoding the input instance and decoding it conditioned on the value of the attribute of interest $v'$.

\subsubsection{Multiple Attribute Changes:} 
\label{subsec:multiple-attribute-changes}
TabChange extends to changing multiple attributes simultaneously.
Let $\mathcal{A} = \{a_1, a_2, \ldots, a_k\}$ denote the set of attributes to be changed. If all attributes in $\mathcal{A}$ have a weak relationship with the other attributes (MI $< \tau$), we directly flip them to the desired values. Otherwise, we employ the adversarial framework with the decoder conditioned on the target values of all attributes in $\mathcal{A}$. The discriminator is trained with separate prediction heads, one for each attribute in $\mathcal{A}$, while the reconstruction loss is computed on the non-conditioned attributes. At inference, we encode the instance and decode it conditioned on the target values to obtain the counterfactual with the necessary changes.

\section{Experimental Design}
\label{sec:experimental-design}
Our evaluation investigates the following research questions:\\
\noindent\textbf{RQ1:} How effective is TabChange compared to baselines for a single attribute?\\
\noindent\textbf{RQ2:} How effective is TabChange compared to baselines for multiple attributes?\\
\noindent\textbf{RQ3:} How efficient is TabChange compared to baselines?

\subsection{Baselines:}
We compare TabChange against two baselines for attribute modification in tabular data.
\subsubsection{ Conditional Variational Autoencoder:} 
CVAE \cite{majumdar2021counterfactuals} generates counterfactuals by conditioning the decoder on a different attribute value than the one used during encoding. The encoder creates a latent representation conditioned on the original attribute value, and the decoder reconstructs the instance conditioned on the target attribute value to produce the counterfactual.
\subsubsection{ Conditional Diffusion Model:}
We implement a diffusion baseline using the official TabDDPM implementation \cite{kotelnikov2023tabddpm}  and train an attribute-conditional diffusion model for each dataset. To the best of our knowledge, existing tabular diffusion models primarily target high-fidelity synthesis and imputation rather than instance-level, identity-preserving, attribute-targeted counterfactual editing. We therefore adapt TabDDPM to enable attribute-conditioned editing of individual instances. For counterfactual generation, we partially noise each input instance to timestep $t$ and then denoise back to $t=0$ while conditioning on the target attribute value. We select optimal timestep $t$ using a validation set by sweeping $t \in \{200, \ldots, 800\}$ and choosing the best trade-off between the successful attribute prediction and proximity to the original instance. This selected value of $t$ is then used for counterfactual generation on the test set.

\subsection{Datasets:}

We evaluate our approach on six real datasets \cite{acs_public_coverage_al_2018} \cite{adult_2} \cite{propublica_compas_analysis} \cite{statlog_(german_credit_data)_144} \cite{bank_marketing_222} \cite{openml_nlsb_43892_v1} commonly used in the tabular domain. Each dataset presents a mixture of categorical and continuous attributes. Additionally, we include a synthetic dataset called the Simple Dataset, which has very high mutual information between the attribute of interest and other attributes, to evaluate our approach. We focus on sensitive attributes (\textit{race}, \textit{age}, \textit{sex}) as the attributes of interest for modification, consistent with the evaluation in \cite{majumdar2021counterfactuals}.

\begin{table}[h]
  \centering
  \caption{Datasets used in our experiments.}
  \label{tab:datasets}
  \resizebox{\linewidth}{!}{%
  \begin{tabular}{llllll}
    \toprule
    Name & Instances & \#Continuous & \#Categorical & Attribute \\
    \midrule
    ACS Public Coverage   & 17,850 & 2 & 16 & sex \\
    Adult                 & 30,160 & 4 & 8  & sex, race \\
    COMPAS                & 6,172  & 5 & 3  & sex, race  \\
    Credit                & 1,000  & 6 & 14 & sex, age \\
    Bank Marketing        & 7,842  & 5 & 11 & age \\
    NLSB                  & 4,908  & 6 & 9  & sex  \\
    Simple Dataset  & 19,998 & 3 & 4 & race, sex  \\
    \bottomrule
  \end{tabular}%
  }
\end{table}

The datasets are obtained from their respective sources, and the instances with the missing attributes are removed. Attributes of \cite{statlog_(german_credit_data)_144} \cite{propublica_compas_analysis} \cite{bank_marketing_222} are preprocessed per \cite{bellamy2019ai}. The numerical attributes are normalized, and the categorical attributes are one-hot-encoded. The dataset follows the standard 80/20 train/test split.

\subsection{Evaluation Metrics}

\subsubsection{Ratio of Instances Retained:}
Counterfactuals are considered natural if they satisfy constraints mined from the dataset. We generate these constraints using the FP-Growth algorithm \cite{han2000mining} independent of the dataset label. These constraints are applied to the generated counterfactuals to evaluate each approach's effectiveness in generating natural instances.
\subsubsection{Distance to the Original Instance:}
A retained counterfactual (one that satisfies dataset constraints) is proximal to its original instance if it has a low Euclidean distance, indicating that minimal changes were made. We measure proximity by computing the Euclidean distance between each counterfactual and its original instance. For each test instance, among all methods that produce natural counterfactuals, the method with the smallest distance is considered to have the most proximal counterfactual.

\subsubsection{Valid Counterfactual Rate:}
The Valid Counterfactual Rate (VCR) is the percentage of test instances for which a method produces a valid counterfactual. This provides a single, unified metric to assess which approach more consistently generates counterfactuals that are both natural and more proximal than competing methods. If all of the methods retain an instance and have the same distance to the original instance, all of the counterfactuals are counted as valid.

\subsubsection{Invalid Counterfactual Rate:}
A counterfactual is considered invalid if it either violates dataset constraints (unnatural) or if another approach generates a counterfactual that was closer to the original instance. Then, the Invalid Counterfactual Rate (ICR) is the percentage of test instances for which an approach generates an invalid counterfactual.

\section{Results and Discussion:}
In this section, we evaluate the counterfactual generated by our proposed approach, TabChange, against the CVAE and Diffusion baseline. Our evaluation is guided by the research questions defined in Section~\ref{sec:experimental-design}, with the primary metrics being the VCR and ICR, which assess both the naturalness and proximity of the generated instances.

\subsection{Mutual Information}
The mutual information between the attribute of interest and other attributes determines whether TabChange applies simply flipping or the adversarial framework. Table~\ref{tab:mi_by_dataset} shows the top mutual information values for each dataset, with values for attributes of interest highlighted in bold along with their ranking position.
We selected an MI threshold of 0.1 based on prior works \cite{murphy2024information} \cite{battiti1994using} \cite{jack2014information}. This threshold effectively separates weak and strong attribute relationships across our datasets. For instance, the COMPAS dataset exhibits very weak attribute relationships (MI $\leq$ 0.05), while the Simple Dataset—designed specifically to evaluate our approach under strong dependencies—shows high MI values (up to 1.10 between race and hw\_k).

The high mutual information values in the Simple Dataset indicate strong attribute dependencies, meaning that simply flipping the attribute of interest without adjusting correlated attributes would generate unnatural counterfactuals. When race or sex is simply flipped in this dataset, nearly all counterfactuals violate dataset constraints. Conversely, for datasets like COMPAS with weak attribute relationships, simple flipping suffices to generate valid counterfactuals.

\begin{table}[t]
  \centering
  \caption{ Mutual information (MI) values by dataset, along with the indices of their position in the dataset in decreasing hierarchy of MI. MI values are in bold for the attributes of interest for that dataset.}
  \label{tab:mi_by_dataset}

  \begingroup
  \begin{tabular}{@{} l c l l c @{}}
    \toprule
    Dataset & Index & Attribute 1 & Attribute 2 & MI Value \\
    \midrule
    \multirow{3}{*}{\makecell[l]{ACS Public\\Coverage}}
      & 1  & sex & fer & \textbf{0.31} \\
      & 2  & pincp & esr & 0.3 \\
      & 3 & age & mar & 0.29 \\
    \midrule
    \multirow{4}{*}{Adult}
      & 1  & marital-status & relationship   & 0.73 \\
      & 2  & relationship   & sex            & \textbf{0.27} \\
      & 6  & marital-status   & sex            & \textbf{0.12} \\
      & 9 & race           & native-country & \textbf{0.1} \\
    \midrule
    \multirow{3}{*}{\makecell[l]{Bank\\Marketing}}
      & 1   & day\_of\_week & month   & 0.73 \\
      & 2   & month         & pdays   & 0.71 \\
      & 9 & age           & job & \textbf{0.11} \\
    \midrule
    \multirow{4}{*}{COMPAS}
      & 1 & age     & priors\_count         & 0.05 \\
      & 5 & age             & race        & \textbf{0.02} \\
      & 11 & sex              & priors\_count         & \textbf{0.01} \\
    \midrule
    \multirow{3}{*}{Credit}
      & 1  & Attribute2  & Attribute5 & 0.35 \\
      & 21  & sex  & Attribute11 & \textbf{0.04} \\
      & 23 & age  & Attribute15 & \textbf{0.04} \\
    \midrule
    \multirow{2}{*}{NLSB}
      & 1 & sex  & height      & \textbf{0.32} \\
      & 4 & sex  & weight & \textbf{0.22} \\
    \midrule
    \multirow{3}{*}{\makecell[l]{Simple\\Dataset}}
      & 1  & race & hw\_k & \textbf{1.10} \\
      & 2  & sex  & h\_cm    & \textbf{0.69} \\
      & 3  & sex & yc    & \textbf{0.69} \\
    \bottomrule
  \end{tabular}
  \endgroup
\end{table}

% ---------------- Single-attribute: Retained + Distance ----------------
\begin{table*}[t]
\centering
\caption{Single-attribute results: ratio of instances retained (\%) and mean distance (bold: higher retained, lower distance).}
\label{tab:single-ret-dist}
\small
\setlength{\tabcolsep}{4pt}
\renewcommand{\arraystretch}{1.05}

\begin{tabular}{llrrrrrr}
\toprule
 &  & \multicolumn{2}{c}{CVAE} & \multicolumn{2}{c}{Diffusion} & \multicolumn{2}{c}{TabChange} \\
\cmidrule(lr){3-4}\cmidrule(lr){5-6}\cmidrule(lr){7-8}
Dataset & Attribute & Retained & Dist. & Retained & Dist. & Retained & Dist. \\
\midrule

ACS Public
Coverage & sex & \textbf{99.9} & \textbf{2.34} & 76.8 & 3.87 & 99.1 & 4.52 \\
Adult & race & 99.9 & 3.26 & 98.4 & 2.18 & \textbf{100} & \textbf{1.41} \\
Adult & sex & \textbf{100} & 3.47 & 82.8 & 3.87 & 99.8 & \textbf{2.75} \\
Bank Marketing & age & \textbf{97.8} & 5.35 & 93.8 & \textbf{2.67} & 96.7 & 6.66 \\
COMPAS & race & 99.7 & 2.36 & 99.8 & 2.09 & \textbf{100} & \textbf{1.41} \\
COMPAS & sex & 99.6 & 2.28 & \textbf{99.9} & 2.37 & 99.4 & \textbf{1.41} \\
Credit & age & \textbf{79} & 8.08 & 45.5 & 3.45 & 43 & \textbf{1.41} \\
Credit & sex & \textbf{85} & 7.47 & 46 & 3.82 & 45 & \textbf{2} \\
NLSB & sex & \textbf{100} & 4.23 & \textbf{100} & 5.73 & \textbf{100} & \textbf{3.33} \\
Simple Dataset & race & \textbf{99.6} & 2.26 & 65.2 & 2.24 & 98.4 & \textbf{2.18} \\
Simple Dataset & sex & \textbf{99.4} & 2.76 & 42.2 & \textbf{2.26} & 97.7 & 2.64 \\
\bottomrule
\end{tabular}
\end{table*}

\begin{table*}[t]
\centering
\caption{Single-attribute results: VCR (TP rate) and ICR (FP rate) for CVAE, Diffusion, and TabChange (bold: higher VCR, lower ICR).}
\label{tab:single-vcr-icr}
\small
\setlength{\tabcolsep}{3.0pt}
\renewcommand{\arraystretch}{1.05}

\begin{tabular}{llr rrrrrr}
\toprule
 &  &  & \multicolumn{2}{c}{CVAE (\%)} & \multicolumn{2}{c}{Diffusion (\%)} & \multicolumn{2}{c}{TabChange (\%)} \\
\cmidrule(lr){4-5}\cmidrule(lr){6-7}\cmidrule(lr){8-9}
Dataset & Attribute & Total & VCR & ICR & VCR & ICR & VCR & ICR \\
\midrule
ACS Public Coverage & sex & 3570 & \textbf{97.4} & \textbf{2.6} & 1.3 & 98.7 & 1.5 & 98.5 \\
Adult & race & 6032 & 0.7 & 99.3 & 1.7 & 98.3 & \textbf{100} & \textbf{0} \\
Adult & sex & 6032 & 32.8 & 67.2 & 4.8 & 95.2 & \textbf{63.9} & \textbf{36.1} \\
Bank Marketing & age & 1569 & 11 & 89 & \textbf{85.3} & \textbf{14.7} & 3.8 & 96.2 \\
COMPAS & race & 1235 & 12.2 & 87.8 & 3.8 & 96.2 & \textbf{100} & \textbf{0} \\
COMPAS & sex & 1235 & 12.2 & 87.8 & 3.8 & 96.2 & \textbf{99.4} & \textbf{0.6} \\
Credit & age & 200 & 32 & 68 & 10.5 & 89.5 & \textbf{43} & \textbf{57} \\
Credit & sex & 200 & 37.5 & 62.5 & 7.5 & 92.5 & \textbf{45} & \textbf{55} \\
NLSB & sex & 982 & 28.6 & 71.4 & 2.3 & 97.7 & \textbf{69.5} & \textbf{30.5} \\
Simple Dataset & race & 4002 & \textbf{49.5} & \textbf{50.5} & 26.1 & 73.9 & 26.8 & 73.2 \\
Simple Dataset & sex & 4002 & 32.2 & 67.8 & 24.7 & 75.3 & \textbf{44.1} & \textbf{55.9} \\
\bottomrule
\end{tabular}
\end{table*}

\subsubsection*{\textbf{RQ1: How effective is TabChange compared to baseline for a single attribute?}}
\leavevmode\par\noindent
Table \ref{tab:single-ret-dist} reports the retention rate and mean distance of the counterfactuals from original instances for all approaches.
CVAE achieves the highest retention rates (100\% for adult-sex, 99.9\% for ACS Public Coverage-sex), with TabChange achieving comparable retention in most cases (99.8\% for adult-sex, 99.1\% for ACS Public Coverage-sex). The Diffusion baseline shows significantly lower retention across datasets (82.8\% for adult-sex, 76.8\% for ACS Public Coverage-sex), as the partial noising and denoising process introduces variations that violate dataset constraints.

For datasets with weak attribute relationships (Adult-Race, COMPAS), TabChange defaults to simple attribute flipping, achieving 100\% retention with minimal distance (1.41). CVAE's decoder learns to map a continuous latent distribution to instances, providing flexibility to satisfy data constraints even with imperfect latent representations. TabChange's adversarial framework uses a deterministic, attribute-invariant latent space, which can occasionally lead to constraint violations when combined with the new attribute value, resulting in slightly lower retention for some datasets.

TabChange achieves superior proximity in 8 out of 11 cases (73\%). For adult-sex and NLSB-sex, TabChange's distances (2.75, 3.33) are significantly lower than CVAE's (3.47, 4.23), while both achieve comparable proximity for Simple Dataset-sex (2.64 vs 2.76). This advantage comes from TabChange's adversarial framework, which removes attribute information from the latent space before conditioning, enabling targeted adjustments. In contrast, CVAE conditions the encoder on the original attribute value but decodes with a different value, resulting in excessive changes.

Table \ref{tab:single-vcr-icr} reports VCR and ICR for all the approaches.
TabChange achieves higher VCR in 8 out of 11 cases. While CVAE performs better on ACS Public Coverage-sex (97.4\% vs 1.5\%) and Simple Dataset-race (49.5\% vs 26.8\%), TabChange achieves substantial VCR improvements for COMPAS-race/sex (100\%, 99.4\% vs 12.2\%, 12.2\%), Adult-sex (63.9\% vs 32.8\%), and NLSB-sex (69.5\% vs 28.6\%) through superior proximity. For simply flipped attributes, the minimal distance (1.41) ensures high VCR, while the adversarial framework's controlled changes result in lower distances and higher VCR for attributes with high MI.

The Diffusion model struggles to balance attribute change and proximity preservation despite tuning the noising timestep $t$, achieving competitive VCR only on Bank Marketing-age (85.3\%) while performing poorly on other datasets.

% ---------------- Multi-attribute: Retained + Distance ----------------
\begin{table*}[t]
\centering
\caption{Multi-attribute results: ratio of instances retained (\%) and mean distance (bold: higher retained, lower distance).}
\small
\setlength{\tabcolsep}{4pt}
\renewcommand{\arraystretch}{1.05}

\begin{tabular}{llrrrrrr}
\toprule
 &  & \multicolumn{2}{c}{CVAE} & \multicolumn{2}{c}{Diffusion} & \multicolumn{2}{c}{TabChange} \\
\cmidrule(lr){3-4}\cmidrule(lr){5-6}\cmidrule(lr){7-8}
Dataset & Attributes & Retained & Dist. & Retained & Dist. & Retained & Dist. \\
\midrule
Adult & race,sex & \textbf{99.9} & 3.74 & 62 & \textbf{2.62} & \textbf{99.9} & 3.13 \\
COMPAS & sex,race & \textbf{99.5} & 2.82 & 98.6 & 2.46 & 97.2 & \textbf{2} \\
Simple Dataset & sex,race & \textbf{99.7} & 3.58 & 84.3 & 3.94 & 94.6 & \textbf{3.41} \\
Credit & sex,age & \textbf{85} & 7.47 & 46 & 3.82 & 45 & \textbf{2} \\
\bottomrule
\end{tabular}
\label{tab:multi-ret-dist}
\end{table*}

% ---------------- Multi-attribute: VCR + ICR (TP/FP) ----------------
\begin{table*}[t]
\centering
\caption{Multi-attribute results: VCR (TP rate) and ICR (FP rate) for CVAE, Diffusion, and TabChange (bold: higher VCR, lower ICR).}
\label{tab:multi-vcr-icr}
\small
\setlength{\tabcolsep}{3.0pt}
\renewcommand{\arraystretch}{1.05}
\begin{tabular}{llr rrrrrr}
\toprule
 &  &  & \multicolumn{2}{c}{CVAE (\%)} & \multicolumn{2}{c}{Diffusion (\%)} & \multicolumn{2}{c}{TabChange (\%)} \\
\cmidrule(lr){4-5}\cmidrule(lr){6-7}\cmidrule(lr){8-9}
Dataset & Attributes & Total & VCR & ICR & VCR & ICR & VCR & ICR \\
\midrule
Adult & race,sex & 6032 & 25.8 & 74.2 & \textbf{45.8} & \textbf{54.2} & 30 & 70 \\
COMPAS & sex,race & 1235 & 35.2 & 64.8 & 14.3 & 85.7 & \textbf{97.2} & \textbf{2.8} \\
Simple Dataset & sex,race & 4002 & 37 & 63 & 16.1 & 83.9 & \textbf{48.1} & \textbf{51.9} \\
Credit & sex,age & 200 & 37.5 & 62.5 & 7.5 & 92.5 & \textbf{45} & \textbf{55} \\
\bottomrule
\end{tabular}
\end{table*}

% --- Single-attribute table ---
\begin{table}[t]
\centering
\caption{Time comparison for single-attribute change (in mins; bold: lowest time taken).}
\label{tab:time-single-3methods}
\small
\setlength{\tabcolsep}{4pt}
\renewcommand{\arraystretch}{1.05}

\begin{tabular}{llrrr}
\toprule
Dataset & Attribute & CVAE & Diffusion & TabChange \\
\midrule
ACS Public Coverage & sex  & 8.18  & \textbf{0.91} & 32.25 \\
Adult                & race & 17.78 & 0.64          & \textbf{0.00} \\
Adult                & sex  & 17.51 & \textbf{0.68} & 46.37 \\
Bank Marketing      & age  & 11.75 & \textbf{0.70} & 23.48 \\
COMPAS               & race & 23.31 & 0.52          & \textbf{0.00} \\
COMPAS               & sex  & 22.64 & 0.52          & \textbf{0.00} \\
Credit               & age  & 10.37 & 0.76          & \textbf{0.00} \\
Credit               & sex  & 10.44 & 0.75          & \textbf{0.00} \\
NLSB                 & sex  & 13.84 & \textbf{0.68} & 29.95 \\
Simple Dataset & race & 36.30 & \textbf{0.53} & 62.65 \\
Simple Dataset & sex  & 36.23 & \textbf{0.58} & 57.12 \\
\bottomrule
\end{tabular}
\end{table}

\begin{table}[t]
\centering
\caption{Time comparison for multiple-attribute change (in mins; bold: lowest time taken).}
\label{tab:time-multi-3methods}
\small
\setlength{\tabcolsep}{4pt}
\renewcommand{\arraystretch}{1.05}

\begin{tabular}{llrrr}
\toprule
Dataset & Attributes & CVAE & Diffusion & TabChange \\
\midrule
Adult   & race, sex & 18.02 & \textbf{0.64} & 46.53 \\
COMPAS  & sex, race & 18.45 & 0.42          & \textbf{0.00} \\
Credit  & sex, age  & 11.10 & 0.66          & \textbf{0.00} \\
Simple Dataset & sex, race & 22.35 & \textbf{0.54} & 39.42 \\
\bottomrule
\end{tabular}
\end{table}

\subsubsection*{\textbf{ RQ2: How effective is TabChange compared to baseline for multiple attributes?}}
\leavevmode\par\noindent
Table \ref{tab:multi-ret-dist} reports the retention rate and mean distance of the counterfactuals for multiple attributes from original instances for all approaches.
CVAE achieves the highest retention (99.9\% for Adult, 99.5\% for COMPAS, 99.7\% for Simple Dataset), with TabChange showing slightly lesser retention (99.9\%, 97.2\%, 94.6\%) but lower retention on Credit (45\%). Diffusion shows substantial retention drops across datasets (62\%, 98.6\%, 84.3\%, 46\%).

TabChange achieves superior proximity in 3 out of 4 cases (75\%). For COMPAS and Credit, where both attributes have weak MI, TabChange's direct flip yields the minimum distance (2.0), substantially lower than CVAE (2.82, 7.47) and Diffusion (2.46, 3.82). For the Adult and Simple Dataset, TabChange achieves better proximity than CVAE, respectively (3.13 vs 3.74, 3.41 vs 3.58). The Diffusion baseline's validation procedure selects t=200, a low noise level that makes minimal changes but violates dataset constraints, resulting in high retention drops.

Table \ref{tab:multi-vcr-icr} reports VCR and ICR for multiple attributes. TabChange achieves higher VCR than CVAE in all 4 cases, with notable improvements on COMPAS (97.2\% vs 35.2\%) and Simple Dataset (48.1\% vs 37\%).

\subsubsection*{\textbf{ RQ3: How efficient is TabChange compared to baselines?}}

Table ~\ref{tab:time-single-3methods} shows the total time (training and inference ) for single-attribute counterfactual generation. For attributes with weak MI, such as race on the Adult and COMPAS and Credit, TabChange directly flips the attribute instantaneously (0.0 mins), whereas CVAE requires a full training cycle (10-23 mins), and diffusion requires 0.5-0.8 mins. 

For attributes with high MI, TabChange uses the adversarial framework, which requires more training time than CVAE and Diffusion methods for training. This is because both the encoder-decoder and discriminator are updated per batch. Diffusion remains fastest (0.5-0.9 mins) but comes at the cost of requiring a separate validation sweep per dataset-attribute combination to select the optimal timestep. Inference time is negligible (a few seconds) for all approaches.

Table ~\ref{tab:time-multi-3methods} shows the timing for multi-attribute training and inference. For datasets where TabChange uses simple flipping (COMPAS, Credit), time is negligible (0.0 mins). For datasets requiring the adversarial framework (Adult, Simple Dataset), baselines outperform TabChange in training efficiency (18-22 mins vs 39-47 mins). However, inference remains instantaneous for all approaches once trained.

\section{Conclusion and Future Work}

Changing attribute values in tabular data requires balancing naturalness (satisfying data distribution constraints) and proximity (minimal change from the original instance). We introduced TabChange, which addresses this challenge through an adaptive, mutual information-guided approach that selects between simple attribute flipping and adversarial training based on attribute relationship strength.

Our evaluation across seven datasets demonstrates that TabChange achieves higher valid counterfactual rates than existing baselines. The key insight is that using a sophisticated generative model is not always necessary—when attributes have weak dependencies, simple flipping suffices, while strong dependencies require the removal of attribute information from the latent space before conditioning.

However, this adaptivity comes with tradeoffs. The adversarial training incurs higher computational costs during the one-time training phase, and the current framework is designed for categorical attributes of interest. Future work could address these limitations by exploring information-theoretic invariance methods for improving training efficiency, extending to continuous attributes by defining meaningful edit magnitudes, and incorporating user-specified constraints on which attributes to modify versus constrain during generation.

\bibliographystyle{ACM-Reference-Format}
\bibliography{tab_change_refs}

\end{document}